\documentclass[conference]{IEEEtran}
\IEEEoverridecommandlockouts
\usepackage{cite}
\usepackage{overpic}
\usepackage{amsmath,amssymb,amsfonts}
\usepackage{algorithmic}
\usepackage{graphicx}
\usepackage[table,xcdraw]{xcolor}
\usepackage{booktabs}
\usepackage{subfigure}
\usepackage{textcomp}
\usepackage[noblocks]{authblk}
\usepackage{xcolor}
\usepackage{tikz}
\def\BibTeX{{\rm B\kern-.05em{\sc i\kern-.025em b}\kern-.08em
    T\kern-.1667em\lower.7ex\hbox{E}\kern-.125emX}}
\newcommand\copyrighttext{%
  \footnotesize \textcopyright 2019 IEEE.  Personal use of this material is permitted.  Permission from IEEE must be obtained for all other uses, in any current or future media, including reprinting/republishing this material for advertising or promotional purposes, creating new collective works, for resale or redistribution to servers or lists, or reuse of any copyrighted component of this work in other works.
  }
\newcommand\copyrightnotice{%
\begin{tikzpicture}[remember picture,overlay]
\node[anchor=south,yshift=10pt] at (current page.south) {\fbox{\parbox{\dimexpr\textwidth-\fboxsep-\fboxrule\relax}{\copyrighttext}}};
\end{tikzpicture}%
}

\begin{document}

\title{\LARGE \bf Visual Semantic SLAM with Landmarks for Large-Scale Outdoor Environment
\thanks{The code of this project has been opened in GitHub: {\tt https://github.com/1989Ryan/Semantic\_SLAM/}}
}

\author[a]{Zirui Zhao}
\author[a]{Yijun Mao}
\author[b]{Yan Ding}
\author[b]{Pengju Ren}
\author[b]{Nanning Zheng}
\affil[a]{Faculty of Electronic and Information Engineering, Xi'an Jiaotong University, Xi'an, China.}
\affil[b]{College of Artificial Intelligence, Xi'an Jiaotong University, Xi'an, China.}


\maketitle
\copyrightnotice
\thispagestyle{empty}
\pagestyle{empty}

\begin{abstract}
Semantic SLAM is an important field in autonomous driving and intelligent agents, which can enable robots to achieve high-level navigation tasks, obtain simple cognition or reasoning ability and achieve language-based human-robot-interaction.  In this paper, we built a system to creat a semantic 3D map by combining 3D point cloud from ORB SLAM\cite{mur2015orb,murORB2} with semantic segmentation information from Convolutional Neural Network model PSPNet-101 \cite{zhao2017pyramid} for large-scale environments. Besides, a new dataset for KITTI\cite{geiger2013vision} sequences has been built, which contains the GPS information and labels of landmarks from Google Map in related streets of the sequences. Moreover, we find a way to associate the real-world landmark with point cloud map and built a topological map based on semantic map. 
\end{abstract}


\section{Introduction}

Semantic 3D environments are increasingly important in multiple fields, especially in robotics. Nowadays, 3D mapping methods only contains odometry or geometrical information of surrounding environments without semantic meanings, which cannot enable robots to infer more information for specific tasks and makes it difficult for human-robot interaction. A map with semantic information allows robots to fully understand their environments, and generalize its navigation capability, just as human does, and achieves higher-level tasks. Semantic information will also enable robots to obtain simple cognition or reasoning ability. Robot perception within semantic information also makes it possible for robots to achieve language-based human-robot interaction tasks. 

Semantic Simultaneously Localization and Mapping (SLAM) system mainly involves the 3D mapping and semantic segmentation. Recently, researches on semantic SLAM are mainly focusing on indoor environments or Lidar based SLAM system for outdoor environments. Visual based Semantic SLAM is mainly achieved by using RGB-Depth (RGB-D) camera, which can be greatly affected by lighting conditions and not well-suited for outdoor environments. Lidar is more suitable in such environment, but it is much more costly than camera-based SLAM system. And Lidar contains less information than visual information, which makes the study in camera-based semantic SLAM system more meaningful. 

We were inspired by human visual navigation system. Human navigation system greatly relies on visual perception since the visual images contain considerable information such as odometry, geometrical structures, and semantic meanings. Our navigation from one place to another is mainly based on landmark level semantic meanings, visual features and their topological relationship. In our system, we use features based on Monocular Visual SLAM system-ORB SLAM2\cite{mur2015orb}. This system is performed by using Oriented FAST and Rotated BRIEF (ORB) features\cite{rublee2011orb}, which has good robustness for moving condition and good real-time performances. It can be used in multiple scenes of outdoor environments with great performance in loop closing. We use ORB-SLAM to extract visual features for re-localization. The semantic information is obtained by Deep Neural Network (DNN). We use PSPNet-101 model\cite{zhao2017pyramid} for pixel-level image semantic segmentation with 19 different semantic labels, including vehicles, buildings, vegetation, sidewalks and roads. The semantic information is then associated with the point cloud map at pixel level. With semantic meaning, we associate the building landmarks with semantic point cloud. We associate the landmarks obtained from Google Map with our semantic 3D map for urban area navigation. It can achieve landmark-based re-localization without GPS information. 

The contributions of this paper are summarized as follows:
\begin{itemize}
\item We developed a system to build a semantic 3D map by fusing visual SLAM map with Semantic Segmentation information for large-scale environments.
\item We developed a new dataset for KITTI\cite{geiger2013vision} sequences, containing the GPS information and labels of landmarks from Google Map in related streets of the sequences. 
\item •	We developed a way to associate the real-world landmark with point cloud map and built a topological map based on semantic map. 
\end{itemize}

This paper is organized as follows: Section 2 introduce the related works in semantic segmentation, SLAM and semantic SLAM. Section 3 describes the details of our proposed methodologies. Section 4 describes experiments in KITTI dataset and analyzes our experiments results. Finally, the conclusions are drawn in Section 5.

\section{Related Work}

The goal of semantic SLAM is to construct semantically meaningful maps where the semantic meanings are attached to the entities by combining geometric and semantic information. SLAM is implemented as a method to rebuild the 3D map of an unknown environment and semantic segmentation is used to extract semantic features. 

SLAM systems depend on the input provided by different kinds of sensors for geometric 3D map and simultaneous estimation of the position and orientation. They can be mainly divided into three categories based on the sensors used for localization, i.e. Lidar-based SLAM, odometry directly provided method and visual SLAM.
The first one is Lidar-based SLAM methods. Laser ranging systems are accurate active sensors. Bosse and Zlot\cite{bosse2009continuous} proposed a method to produce locally accurate maps by matching geometric structures of local point clusters using a 2-axis Lidar. Zhang and Singh\cite{zhang2014loam} developed Lidar odometry and mapping (LOAM) approach which estimates odometry and motion of vehicle and produces 3D maps in real-time. However, these methods have trouble to accurately map or localize if there are few structural features in current environments. The second category is to be provided the odometry directly using independent position estimation sensors, e.g. GPS/INS. It is the most commonly applied to build large-scale 3D maps for autonomous vehicle\cite{puente2013review}. Although this method is capable of making improvement in the accuracy of mapping, it often costs a lot due to the expensive sensors and has limitations in indoor applications of mobile robotics\cite{cadena2016past}. 
Many recent researches focus on using
visual information solely, which is specifically referred to as visual SLAM. 
This method has been widely adopted in the field of computer vision, robotics, and AR\cite{azuma1997survey}. 
Davison et al\cite{davison2007monoslam} proposed the first monocular visual SLAM system in 2007, named MonoSLAM, which only uses a monocular camera to estimate 3D trajectory. To solve the problem of the computational cost in MonoSLAM, PTAM\cite{klein2007parallel} was proposed and in 2015. Mur-Artal and Tards proposed ORB-SLAM\cite{mur2015orb, murORB2}, which is one of visual SLAM systems with full sensor support and best performance, with applying ORB features in parallel tracking, mapping, and loop closure detection, and using pose graph optimization and bundle adjustment\cite{triggs1999bundle} based optimization. Another kind of visual SLAM systems, unlike feature-based methods mentioned above, directly uses images as input without any abstraction with descriptors or handcrafted feature detectors, called direct methods\cite{taketomi2017visual}. DTAM\cite{newcombe2011dtam}, in which tracking is implemented by associating the input image with synthetic view images generated from the reconstructed map, and LSD-SLAM\cite{engel2014lsd}, which follows the idea from semidense VO\cite{engel2013semi}, are the leading strategies in direct methods.
DSO\cite{engel2017direct} combines the minimum photometric error model with the joint optimization method of model parameters. 
In this paper, our proposed model mainly based on ORB-SLAM.

Semantic segmentation is another challenging task in computer vision. Motivated by the development of powerful deep neural networks~\cite{krizhevsky2012imagenet,simonyan2014very,szegedy2015going,he2016deep}, 
semantic segmentation achieves tremendous progress inspired by substituting the fully-connected layer in classification for the convolution layer\cite{long2015fully}. Farabet et al.\cite{farabet2012learning} adopted the multi-scale convolutional network to extract multi-scale features from the image pyramid (Laplacian pyramid version of the input image). Couprie et al.\cite{couprie2013indoor} adopted a similar approach to learn multi-scale features with image depth information. In \cite{liu2016multi}, multi-scale patches for object parsing were generated to achieve segmentation and classification for each patch at the same time and aggregates them to infer objects.
As the development of enhancement of feature based methods\cite{garcia2018survey} which extract features at multi-scale, 
Zhao et al.\cite{zhao2017pyramid} proposed pyramid scene parsing network (PSPNet) for semantic segmentation, which allows multi-scale feature ensembling. It concatenates the feature maps with up-sampled output of parallel pooling layers and involves information with different pyramid scales, varying among different sub-regions. This method achieves a practical system for state-of-the-art semantic segmentation and scene parsing including all crucial implementation details.

Semantic mapping provides an abstraction of space and a means for human–robot interaction. According to \cite{kostavelis2015semantic}, our research can be categorized into outdoors interpretation. Multiple methods have been proposed to confront with the challenge of semantic mapping in outdoor environment. The method proposed in \cite{kostavelis2012collision} was the early work utilizing stereo vision and classifying image to separate the traversable and non-traversable scenes with SVM. Furthermore, the algorithm described in \cite{vineet2015incremental} generated an efficient and accurate dense 3D reconstruction with associated semantic labels. Conditional Random Field (CRF) framework was applied to operate on stereo images to estimate labels and annotate the 3D volume. Cheng et al.\cite{sengupta2013urban} applied ORB-SLAM to get real-scale 3D visual maps and CRF-RNN algorithm for semantic segmentation. In \cite{li1611semi}, this challenge was solved by combining the state-of-the-art deep learning algorithms and semi-dense SLAM based on a monocular camera. 2D semantic information are transferred to 3D mapping via correspondence between connective Keyframes with spatial consistency.
However, there are few works about associating the real-world landmark with semantic 3D map for task-based navigation and human-robot interaction.

\section{Approach}

\subsection{System overview}

\begin{figure*}[t!]\centering
   \begin{overpic}[width=0.75\textwidth]{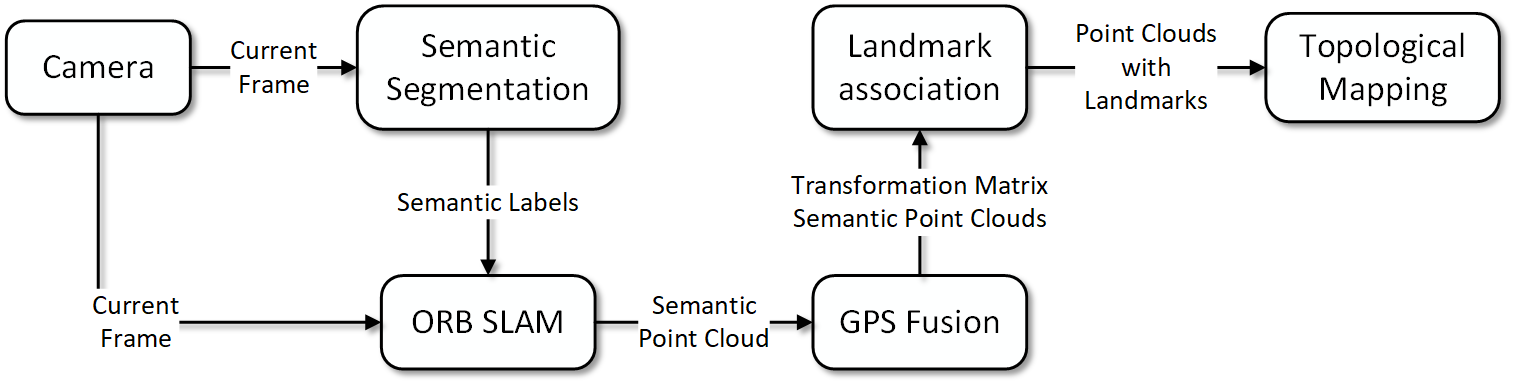}
    \end{overpic}
    \caption{The flowchart of whole system.
    }\label{fig:flowchart}
\end{figure*}

Our Semantic SLAM system uses monocular camera as the main sensor and focuses on large-scale urban areas. As shown in the flowchart, our system can not only reconstruct the 3D environments using ORB feature, but also make it possible for GPS data fusion, map re-utilization and real time re-localization and landmark based localization. The flowchart of whole system is shown in the figure \ref{fig:flowchart}.

 First, the image is segmented by CNN based segmentation algorithm. The pixel-level semantic mapping result and current frame will then be sent to the SLAM system for environment reconstruction. The geometrical environment is reconstructed by ORB SLAM, in which the point cloud is generated by corner ORB features in the current frame. In the SLAM system, the pixel-level semantic information will associate with the map point using Bayesian update rule, which will update probability distribution of each map point for each observation in a frame. Then the landmarks will be projected in the SLAM map and be associated with nearest keyframes saved in SLAM system. The map can be reutilized for landmark-level re-localization without GPS information. We also provide methods to build topological reachable relationship for each landmark, which will be more convenient for robots to achieve landmark-level self-navigation.

\subsection{Semantic mapping}

\subsubsection{Semantic segmentation}

The aim of semantic segmentation is to correctly classify each pixel for their semantic labels. In this work, we choose the PSPNet-101 model\cite{zhao2017pyramid} for image segmentation and TensorRT for real time inference acceleration.


\subsubsection{ORB SLAM2}

The 3D reconstruction is achieved by ORB SLAM \cite{mur2015orb}, an open-source visual-feature-based state-of-the-art SLAM system. ORB SLAM has good real time performance with fantastic loop closing. We use ORB SLAM for 3D reconstruction and trajectory estimation. There are three threads, i.e. tracking, local mapping and loop closing, run parallelly in the ORB SLAM system.

\subsubsection{Real time data fusion}

The data fusion step is trying to associate the semantic meaning with each map point in SLAM system. In this step, we try to use Bayesian update rule to update the probability distribution of semantic label of each map point.

First, the scores over 19 labels at each pixel will be sent to SLAM system. In ORB SLAM system, the good feature point will be saved and transformed in the point cloud. There will be a transform relationship between those feature points in 3D point cloud coordinate system and in camera coordinate system. Transformation relationship between 3D point cloud system and the camera coordinates is shown below:
\begin{equation}\left[
    \begin{array}{c}
          u\\
          v\\
          w
    \end{array}\right]
    = T_{pointcloud2camera}\left[
    \begin{array}{c}
    x_m\\
    y_m\\
    z_m\\
    1
    \end{array}\right]
\end{equation}
\begin{equation}
    \left[
    \begin{array}{c}
    u_c\\
    v_c 
    \end{array}
    \right]=
    \left[
    \begin{array}{c}
    \frac{u}{w}\\
    \frac{v}{w}
    \end{array}
    \right]
\end{equation}
where ($x_m,y_m,z_m$) are the positions of the map point in 3D map coordinates. $T_{pointcloud2camera}$ is the parametric matrix which transfers the position of point cloud to the position in camera coordinates. $(u_c,v_c)$ are camera pixel in camera coordinates  that corresponds to the map point $(x_m,y_m,z_m)$. After the feature point being projected to the camera coordinates, the probability distribution of 19 labels of each feature points will be given as shown below:
\begin{equation}
    L_m (x_m,y_m,z_m )=F_s (u_c,v_c )
\end{equation}
where $F_s$ is the probability distribution of each label in current frame after semantic segmentation section, and $L_m(x_m, y_m, z_m)$ represents the label of the map point in $(x_m, y_m, z_m)$. Moreover, since each feature point can be observed in different frames, data fusion method is applied in different observation. The multi-observation data fusion by using Bayesian update is performed, as shown below:
\begin{equation}
    p(l_l^m|F_{1:k},P_{1:k})=\frac{1}{Z} p(l_k^m|F_k,P_k)p(l_{k-1}^m |F_{1:k},P_{1:k})
\end{equation}
\begin{equation}
    Z=\sum_{m=1}^{19} p(l_k^m |F_k,P_k)p(l_{k-1}^m |F_{1:k},P_{1:k})
\end{equation}
where $Z$ is the normalization constant and $l_k^m$ denotes the labels of map point $m$ at frame $k$. $p(l_l^m|F_{1:k},P_{1:k})$ denotes the cumulative probability distribution from frame 1 to frame $k$, respectively. In this equation, the probability distribution is the result of previous distribution update with newly upcoming frame and point cloud. The probability distribution of each feature point is saved in ORB SLAM system. And the eventual label of each map point is searched by maximizing the probability, which is shown in the equation below. 
\begin{equation}
    L_p (m)= argmax_{l^m} p(l^m|F,P)
\end{equation}
where $m$ denotes a single map point, and $l_m$ represents the semantic label of map point $m$. During the real time fusion, each map point will contain one semantic label and a semantic probability distribution.

\subsection{GPS fusion}
To associate the building landmarks with the point cloud at pixel level to generate the semantic point cloud., we need to convert WGS84 coordinates of building landmarks, which is used in Google map, into the same coordinate system with the point cloud. However, the longitude and latitude in the WGS84 obtained from google map API is not suitable to directly convert. Thus, we first convert the coordinate to Cartesian coordinate, in which the unit is meter. After converting the GPS information the keyframes to Cartesian coordinates, we adopted the method proposed by Besl and McKay\cite{besl1992method} to unify the coordinate system with point cloud. Every 30 frames we took the current frame as the sampling point and added the corresponding pose and the longitude and latitude to the two global samplers. At the end, we used SVD to compute the best rotation between these two point sets in global samplers. Here we assume $P_A$ as the set of points in Cartesian coordinate, $P_B$ as the set of points in pose coordinate. $centroid_A$ is the centroid of $P_A$ and $centroid_B$ is the centroid of $P_B$. As the scales of the two coordinates are different, scale transformation is also required.
The Rotation matrix $R$ and translation matrix $T$ is computed as:
\begin{equation}
    H=\sum_{i=1}^N (P_A^i - centroid_A)(P_B^i - centroid_B)^T
\end{equation}
\begin{equation}
    [U,S,V]=SVD(H)
\end{equation}
\begin{equation}
   R=\frac{1}{\lambda} VU^T
\end{equation}
\begin{equation}
    T=-\frac{1}{\lambda} R\cdot centroid_A + centroid_B
\end{equation}
where $\lambda$ is the scale multiplier since the scale of different coordinates might be different, which is calculated as:
\begin{equation}
    \lambda=average\frac{\lVert P_A - centroid_A\rVert}{\lVert P_B - centroid_B\rVert}
\end{equation}
After obtaining $R$ and $T$, every point in Cartesian coordinate, which was the position of the building landmarks, can be converted to the coordinate system of the point clouds as:
\begin{equation}
    B=R\cdot A+T
\end{equation}
where $A$ is the point in Cartesian coordinate, which represents the position of landmark, and $B$ is the point in point clouds’ coordinate system. Then we can find the corresponding point in point cloud map and fuse semantic label with it.

\begin{figure*}[t!]\centering
   \begin{overpic}[width=\textwidth]{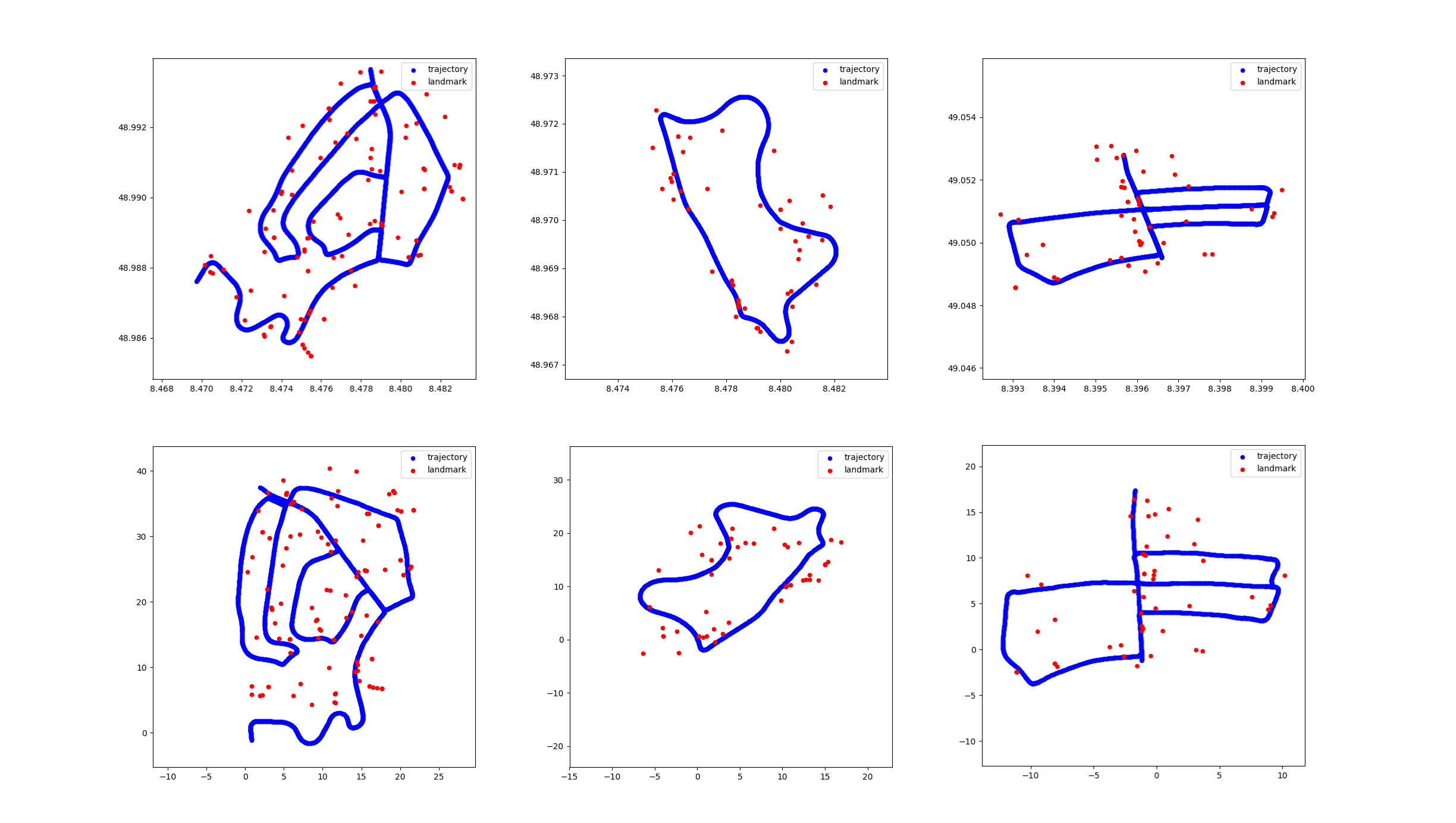}\small
   \put(16,0){(a)~sequence 02}
   \put(45,0){(b)~sequence 09}
   \put(73,0){(c)~sequence 05}
   \end{overpic}
   \caption{ GPS-SLAM transformation result. Figures on the top shows ground truth of GPS positions and figures below shows the transformed positions.
   }\label{fig:gps}
\end{figure*}

\subsection{Post process}

After the real time process, we will perform a post process to optimize the result and get more structured semantic information. In this process, the clustering method will be applied in different semantic labels for object-level semantic map. Landmarks will also be fused in feature points and can be used for landmark-level localization and navigation.

\subsubsection{Landmark level data fusion}

 We assume that the relationship between the landmarks and feature points is fuzzy membership. In one area, the landmark is important for human navigation since the target of human navigation is mainly assigned by landmark. With landmark GPS information and semantic labels, we can make landmark-level data fusion with our 3D reconstruction result, which will be more convenient for task-oriented navigation problem. We will release those datasets with our open-source code. 

We use a fuzzy-mathematics-based method for landmark data fusion. In this method, we will not focus on the accuracy of the location of the landmark, but the membership distribution of the landmark location. Since according to the human cognitive custom, the concept of the location of landmarks are actually a fuzzy concept. This allows the robot to define the position of landmarks in the human's way. We try to evaluate the location membership based on Gaussian probability distribution. If the place is physically near to the landmark, the membership of such place will be higher regarding to the Gaussian distribution. The membership is defined as shown below:
\begin{equation}
    m(x,y)=G(x,y,x_l,y_l,\sigma)
\end{equation}
where the $m(x,y)$ denotes the membership of location at $(x,y)$. $G(x,y,x_l,y_l,\sigma)$ denotes the 2D Gaussian probability density function (PDF). $(x_l,y_l)$ denotes the landmark location, $\sigma$ denotes the standard deviation of the Gaussian distribution. The distribution will be inserted in the semantic map and be associated with the trajectory for real time landmark-based localization. 

\subsubsection{Topological semantic mapping}

The semantic SLAM can also generate a topological semantic map which only contains reachable relationships between landmarks and their geometrical relationships. There will be only edges and nodes in the semantic map and be more suitable for global path planning. 

\begin{figure*}
\centering
\subfigure[Sequence 05]{
\begin{minipage}[t]{0.3\textwidth}
\centering
\includegraphics[width=\textwidth]{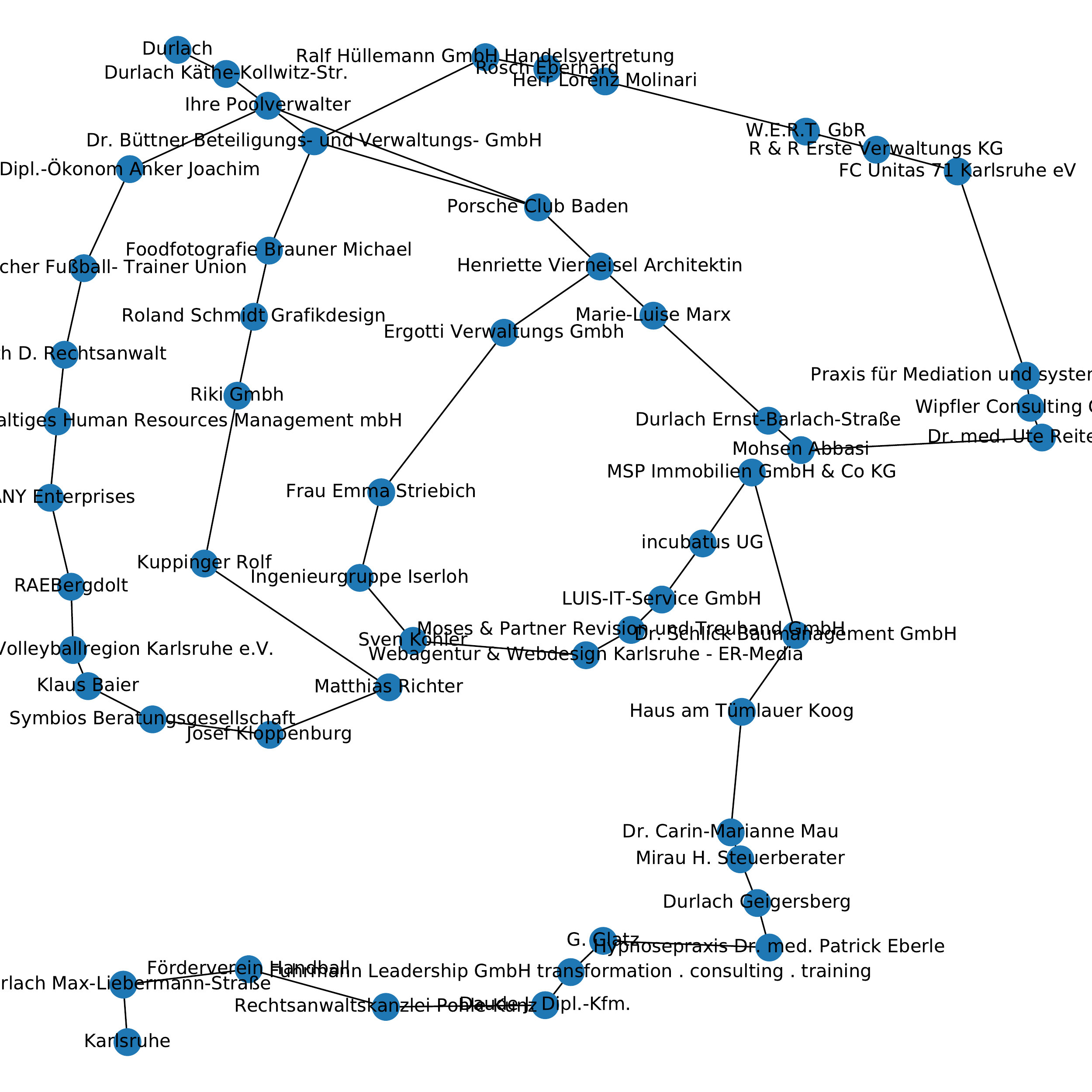}
\end{minipage}
}
\subfigure[Sequence 09]{
\begin{minipage}[t]{0.3\textwidth}
\centering
\includegraphics[width=\textwidth]{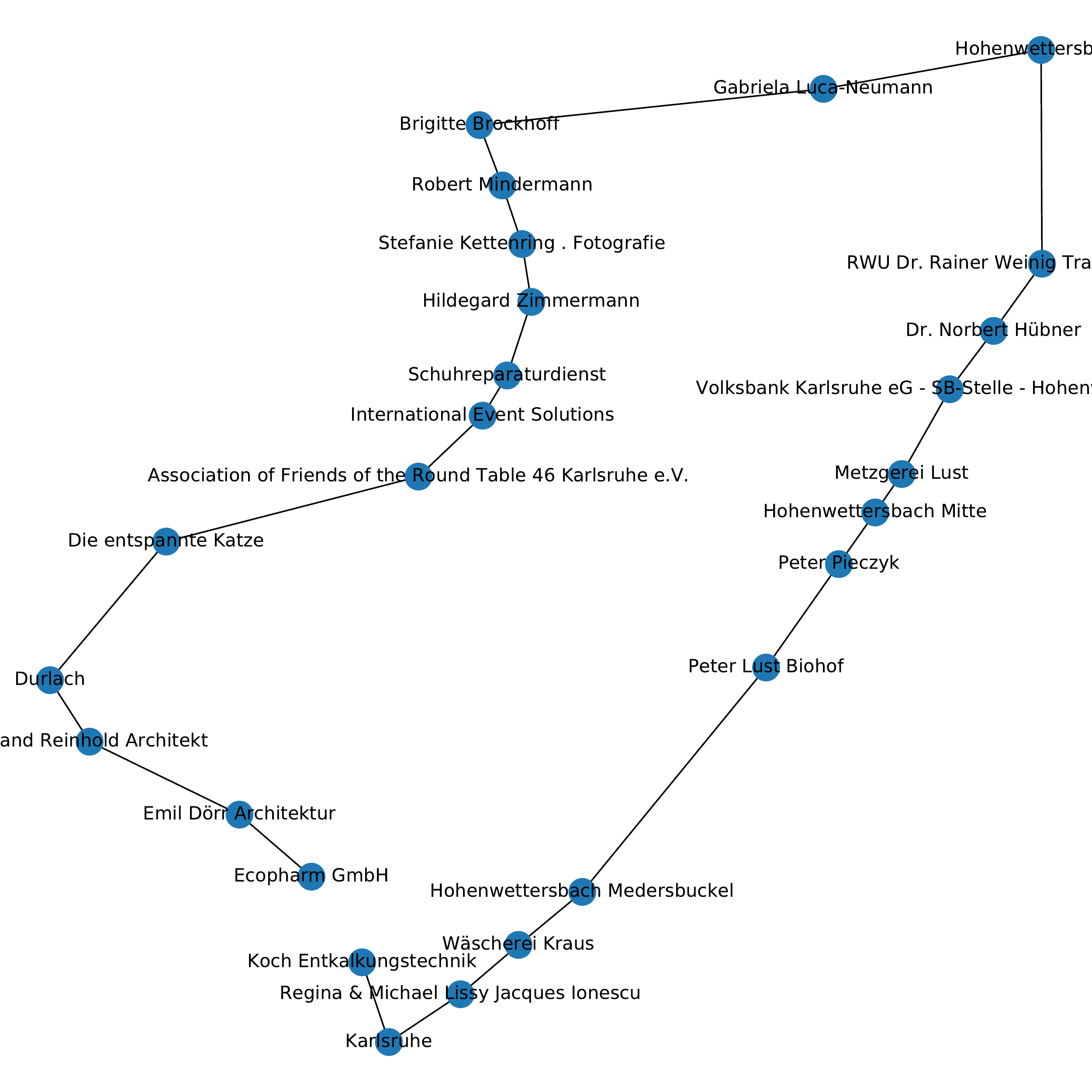}
\end{minipage}
}
\subfigure[Sequence 02]{
\begin{minipage}[t]{0.3\textwidth}
\centering
\includegraphics[width=\textwidth]{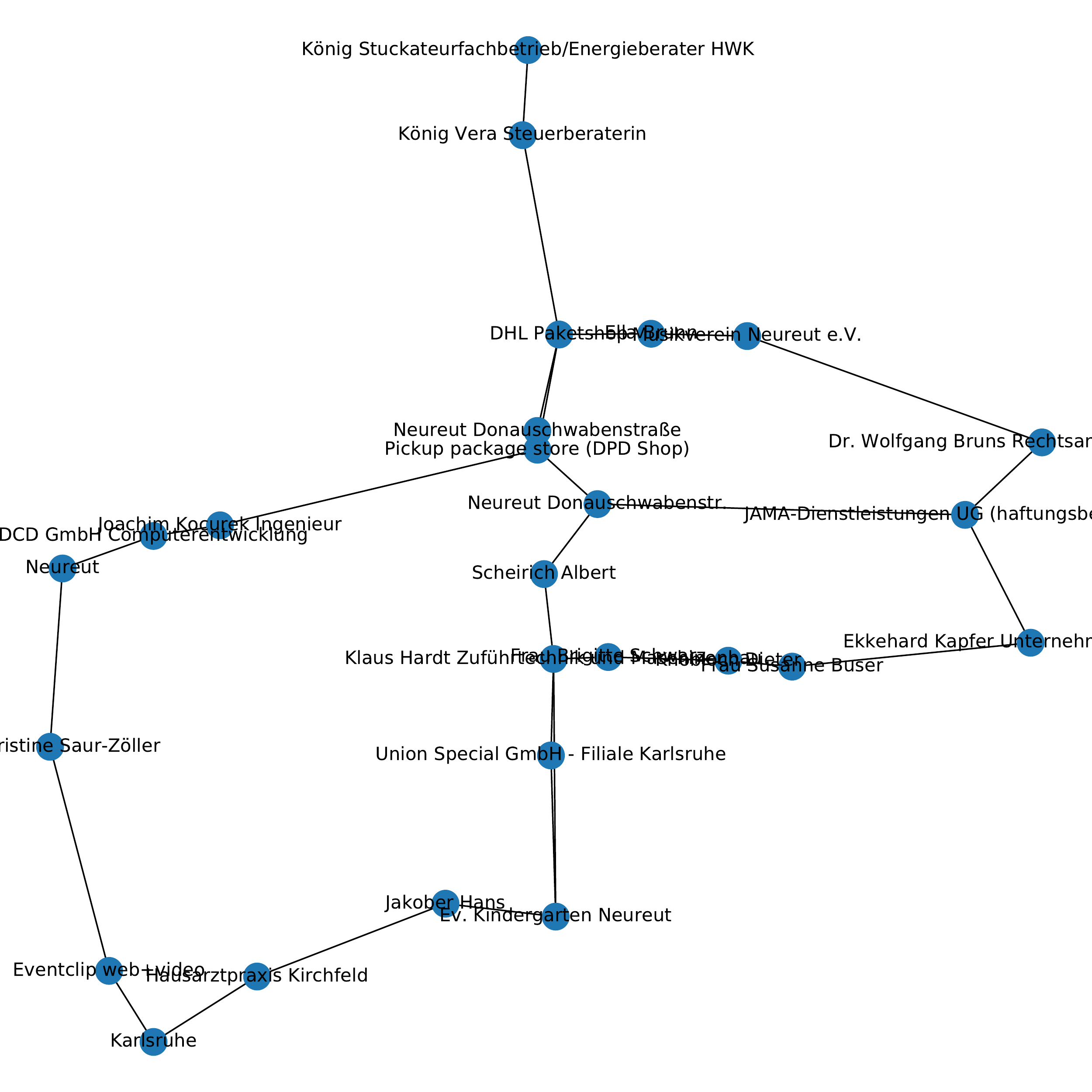}
\end{minipage}
}
\centering
\caption{ Visualization of topological mapping. 
}
\label{fig1}
\end{figure*}

The topological map is built through the following steps. First, after the mapping process in SLAM system, the trajectory of camera will be saved. The landmark will be associated with its closest key frame. Second, there will be two kinds of key frame that are saved, i.e. the key frames associated with landmarks and the key frames in where to turn. Third, the map will be optimized if the place is visited for more than one times. The previous nodes will be fused with the new node if they represent the same location or landmark. The Topological semantic map is shown in the figure \ref{fig1}.

\section{Experiments}

We designed experiments mainly based on the KITTI dataset, which is available to the public and mainly recorded at the urban area. Based on the GPS information recorded in the KITTI raw data, we record the landmark GPS information through Google Map. The dataset contains longitude, latitude and true name of landmarks. We record the sequences 00 to 10 for evaluation and testing. It will be released to the public soon. Besides, we evaluate the quantitative benchmark of the system in real-time performance. The experiments were designed by using ROS and Keras, our computing platform involves Intel Core i7 CPU and NVIDIA GeForce GTX 1080Ti GPU platform. 

\subsection{Dataset}

The KITTI sequences have a large number of outdoor environments at urban areas. We choose sequences 00 to 10 to evaluate the overall quality of our system. The data we use in our system is mainly GPS information and images. We use the RGB images from right camera to simulate the monocular camera. We do not fully rely on GPS information since we just use the GPS information every 30 frames to simulate poor GPS devices in real world implementation.

\subsection{Implementation Details}

First, our experiments are mainly based on Robotic Operating System (ROS), which is a framework for multiple processes communications in robots. We use ROS node to simulate the camera ROS drive and GPS device drive. For all experiments, the transformation relationship between point cloud coordinates and camera coordinates is estimated in ORB SLAM. In GPS fusion and transformation, we use sampling rather than all GPS information to reduce the relative error and simulate the poor GPS signals. We sample the GPS information every 30 frames. Semantic Segmentation is implemented in TensorFlow and Tensor RT. The model is trained in Cityscape datasets.

\subsection{Qualitative Evaluation}

\begin{figure*}
\centering
\subfigure[Sequence 05]{
\begin{minipage}[t]{0.6\textwidth}
\centering
\includegraphics[width=\textwidth]{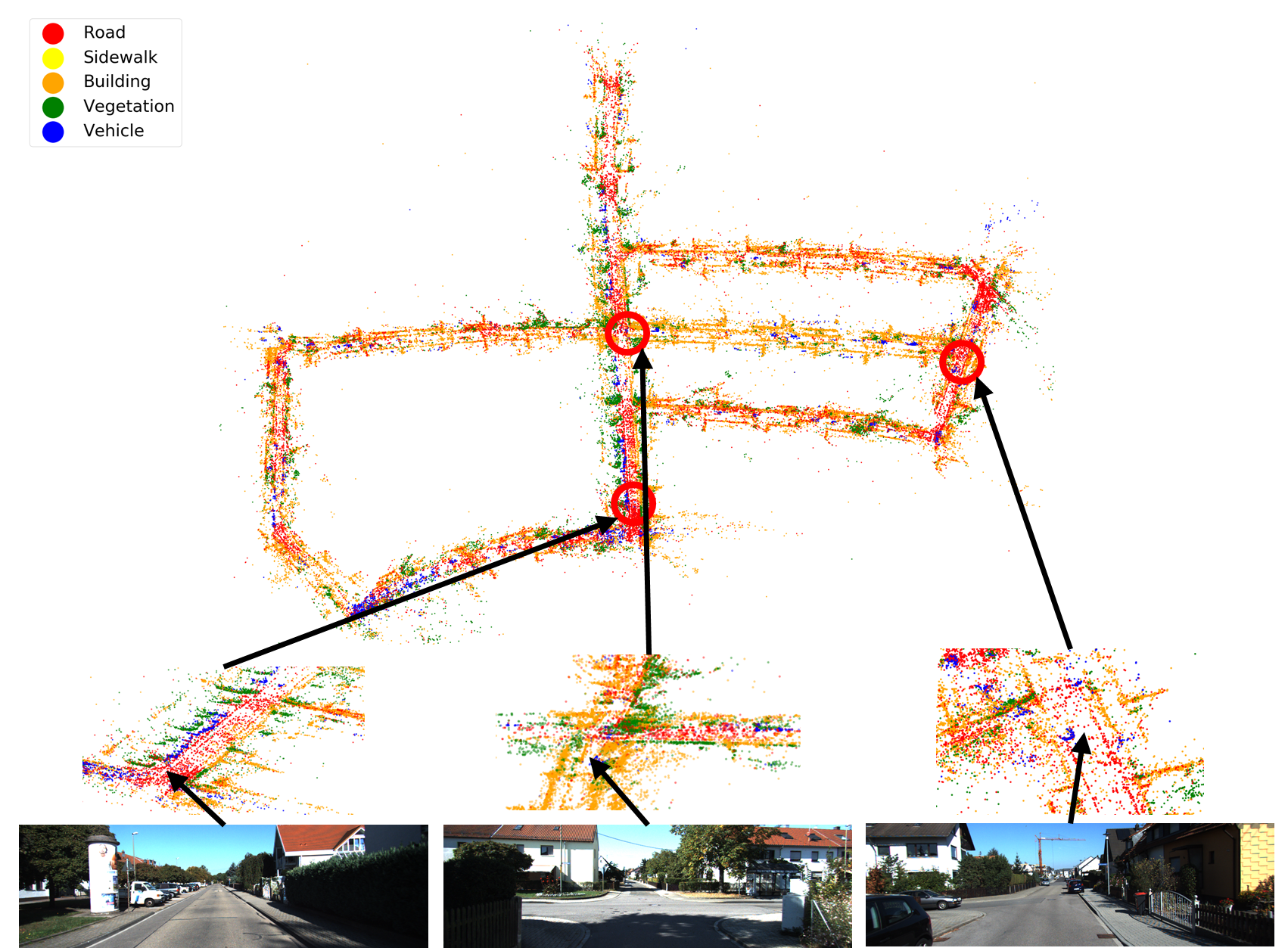}
\end{minipage}
}
\subfigure[Sequence 09]{
\begin{minipage}[t]{0.6\textwidth}
\centering
\includegraphics[width=\textwidth]{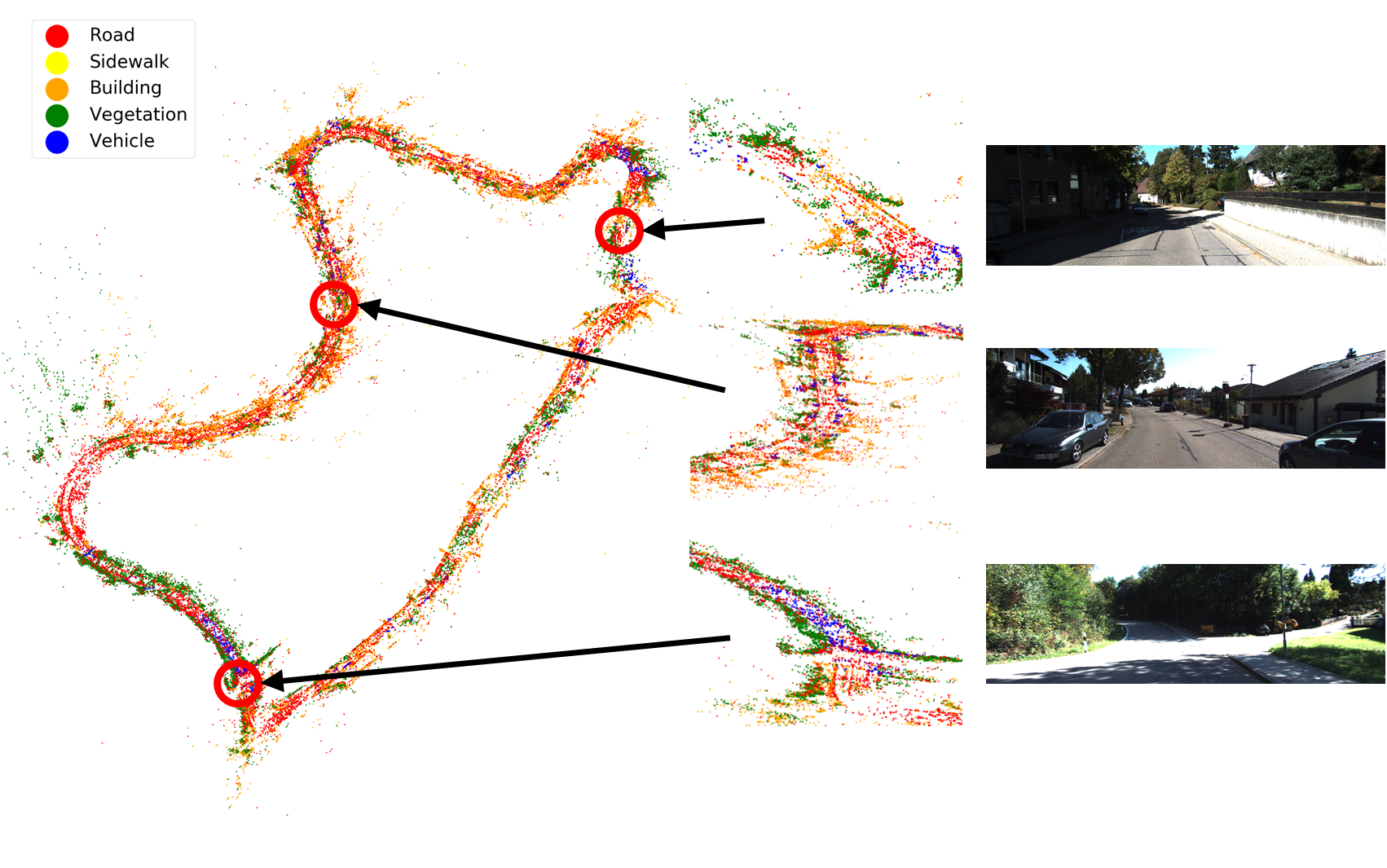}
\end{minipage}
}
\subfigure[Sequence 02]{
\begin{minipage}[t]{0.6\textwidth}
\centering
\includegraphics[width=\textwidth]{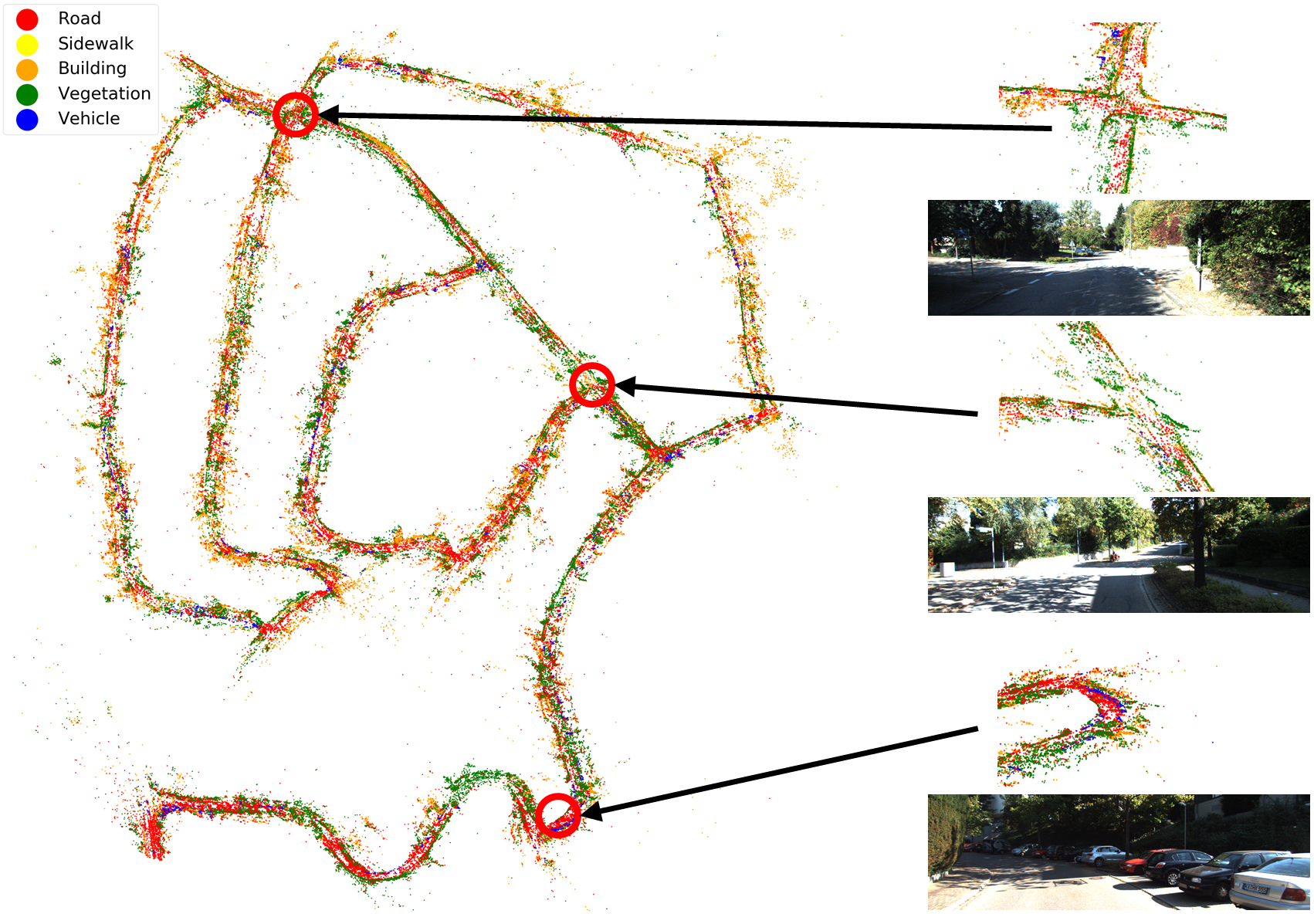}
\end{minipage}
}
\centering
\caption{ Visualization of semantic 3D mapping. Top view of the whole sequences and close-up views of semantic map. 
}
\label{fig:results}
\end{figure*}

In order to evaluate the semantic SLAM system, multiple large-scale outdoor sequences in KITTI datasets were used. The qualitative results of our system are presented in the figure \ref{fig:results}. Each figure shows multiple views of the whole map. The landmark distribution and topological semantic map is also shown in the figure. It shows that our system can successfully fuse the semantic labels into the point cloud generated by ORB SLAM, thereby generating the semantic 3D point cloud with 19 labels. Moreover, the landmark level data fusion is preformed and got good topological relationships in different sequences. It will be useful for large-scale landmark-based navigation tasks or human-robot interaction.

Experiment shows that semantic information will allow the robots to know more about the environments not only the meaningless features but also their semantic meanings. Besides, based on semantic meaning, the robots will re-localize themselves with more robust features such as features on buildings, roads, sidewalks, walls, rather than vehicles, trees, person, etc. We choose sequence 02, 05 and 09 for example. The result is shown in figure \ref{fig:results}.

\subsection{Time Analysis}

The experiments were designed by using ROS and Keras, our computing platform involves Intel Core i7 CPU and NVIDIA GeForce GTX 1080Ti GPU platform. 

We have tested the system run time when they work together. The overall system can run in nearly 1.8Hz in our computing system.  Since the semantic segmentation model we use is based on PSPNet-101 which is a large CNN model without acceleration, we can reach better performance if the model is accelerated in FPGA or TensorRT. The overall run time performance of our system is shown in the table \ref{tab:time_analysis}.

\begin{table}[]
\caption{Time analysis result}
\label{tab:time_analysis}
\centering
\begin{tabular}{c|c}
\hline
Method           & Frequency/Time \\ \hline
PSPNet-101       & 1.8Hz          \\
ORB\_SLAM2       & 15.1Hz         \\
Data Association & 0.0005s        \\ \hline
\end{tabular}
\end{table}

\section{Conclusion}

In this paper, a Monocular camera-based semantic SLAM system with landmarks is developed for large-scale outdoor localization and navigation. Existing works have focused only on accuracy or real-time performance, which might be difficult for real improvement of overall cognitive level of robots. We conducted a dataset based on KITTI GPS information for landmark based semantic fusion and topological semantic mapping. A 3D semantic point cloud with landmark information is built by our system using the dataset we mentioned. It contains real name and position of landmarks, multiple semantic labels, which makes it possible for offline language-based human-robot-interaction, task-oriented navigation or landmark-level localization. The 3D map is fused with related semantic information by using coordinate system transformation and Bayesian update. The landmark data fusion is achieved by fuzzy membership based on Gaussian distribution, by which the topological semantic map is built. 

Our paper provides several compelling fields for future work. We are planning to improve the visual SLAM system to adapt it to localization and navigation in a variety of lighting conditions. Furthermore, we would like to develop a robot navigation system based on landmark topological maps and human-robot-interaction. How to improve the localization performance by using semantic information is also an interesting area worthy of future study.

\section*{Acknowledgment}

This work was supported in part by the National Science and Technology Major Project of China No. 2018ZX01028-101-001 and National Natural Science Foundation of China No.61773307.

\bibliographystyle{./bibliography/IEEEtran}
\bibliography{ref}

\end{document}